\documentclass{article}

\usepackage{arxiv}

\usepackage[utf8]{inputenc} 
\usepackage[T1]{fontenc}    
\usepackage{hyperref}       
\usepackage{url}            
\usepackage{booktabs}       
\usepackage{amsfonts}       
\usepackage{amsmath}
\usepackage{nicefrac}       
\usepackage{microtype}      
\usepackage[capitalise]{cleveref}       
\usepackage{lipsum}         
\usepackage{graphicx}
\usepackage{natbib}
\usepackage{doi}

\usepackage{subcaption} 
\usepackage{bm}

\usepackage{adjustbox}
\usepackage{multicol}
\usepackage{siunitx}
\newrobustcmd\B{\DeclareFontSeriesDefault[rm]{bf}{b}\bfseries}

\usepackage[nolist,nohyperlinks]{acronym}
\newacro{nn}[NN]{neural network}
\newacro{cnn}[CNN]{convolutional neural network}
\newacro{vi}[VI]{variational inference}
\newacro{lrt}[LRT]{local reparameterization trick}
\newacro{elbo}[ELBO]{evidence lower bound}
\newacro{kl}[KL]{Kullback–Leibler}
\newacro{isic}[ISIC]{International Skin Imaging Collaboration}
\newacro{sbd}[SBD]{semantic boundaries dataset}
\newacro{xai}[XAI]{explainable artificial intelligence}

\newcommand{\R}{\mathbb{R}}
\newcommand{\data}{\mathcal{D}}
\newcommand{\normal}{\mathcal{N}} 
\newcommand{\Y}{\mathcal{Y}}
\newcommand{\KL}{D_{\mathrm{KL}}}
\DeclareMathOperator{\EX}{\mathbb{E}}

\DeclareMathOperator*{\argmax}{arg\,max}

\DeclareMathOperator{\vect}{vec}

\newcommand{\ve}{\mathbf{e}}
\newcommand{\vx}{\mathbf{x}}
\newcommand{\vy}{\mathbf{y}}
\newcommand{\vw}{\mathbf{w}}
\newcommand{\va}{\mathbf{a}}
\newcommand{\vb}{\mathbf{b}}
\newcommand{\vtheta}{\bm{\theta}}
\newcommand{\vmu}{\bm{\mu}}
\newcommand{\vsigma}{\bm{\sigma}}
\newcommand{\vepsilon}{\bm{\epsilon}}
\newcommand{\vzero}{\mathbf{0}}

\newcommand{\mE}{\mathbf{E}}
\newcommand{\mA}{\mathbf{A}}
\newcommand{\mB}{\mathbf{B}}
\newcommand{\mW}{\mathbf{W}}
\newcommand{\mdelta}{\bm{\delta}}
\newcommand{\mgamma}{\bm{\gamma}}

\newcommand{\tX}{\mathsf{X}}

\DeclareFontFamily{U}{matha}{\hyphenchar\font45}
\DeclareFontShape{U}{matha}{m}{n}{
      <5> <6> <7> <8> <9> <10> gen * matha
      <10.95> matha10 <12> <14.4> <17.28> <20.74> <24.88> matha12
      }{}
\DeclareSymbolFont{matha}{U}{matha}{m}{n}
\DeclareMathSymbol{\Lt}{3}{matha}{"CE}

\title{Regularizing Explanations in Bayesian Convolutional Neural Networks}

\date{}

\newif\ifuniqueAffiliation

\ifuniqueAffiliation 
\author{{\hspace{1mm}David S.~Hippocampus}\thanks{Use footnote for providing further
		information about author (webpage, alternative
		address)---\emph{not} for acknowledging funding agencies.} \\
	Department of Computer Science\\
	Cranberry-Lemon University\\
	Pittsburgh, PA 15213 \\
	\texttt{hippo@cs.cranberry-lemon.edu} \\
	\And
	{\hspace{1mm}Elias D.~Striatum} \\
	Department of Electrical Engineering\\
	Mount-Sheikh University\\
	Santa Narimana, Levand \\
	\texttt{stariate@ee.mount-sheikh.edu} \\
}
\else
\usepackage{authblk}

\author{%
	{\hspace{1mm}Yanzhe Bekkemoen\thanks{\texttt{yanzhe.bekkemoen@ntnu.no}}}%
}
\author{%
	{\hspace{1mm}Helge Langseth\thanks{\texttt{helge.langseth@ntnu.no}}}%
}
\affil{Department of Computer Science, Norwegian University of Science and Technology, Trondheim, Norway}
\fi

\hypersetup{
pdftitle={Regularizing Explanations in Bayesian Convolutional Neural Networks},
pdfkeywords={Explainable Artificial Intelligence, Deep Learning, Bayesian Neural Networks},
}

\begin{document}
\maketitle

\begin{abstract}
Neural networks are powerful function approximators with tremendous potential in learning complex distributions. However, they are prone to overfitting on spurious patterns. Bayesian inference provides a principled way to regularize neural networks and give well-calibrated uncertainty estimates. It allows us to specify prior knowledge on weights. However, specifying domain knowledge via distributions over weights is infeasible. Furthermore, it is unable to correct models when they focus on spurious or irrelevant features. New methods within explainable artificial intelligence allow us to regularize explanations in the form of feature importance to add domain knowledge and correct the models' focus. Nevertheless, they are incompatible with Bayesian neural networks, as they require us to modify the loss function. We propose a new explanation regularization method that is compatible with Bayesian inference. Consequently, we can quantify uncertainty and, at the same time, have correct explanations. We test our method using four different datasets. The results show that our method improves predictive performance when models overfit on spurious features or are uncertain of which features to focus on. Moreover, our method performs better than augmenting training data with samples where spurious features are removed through masking. We provide code, data, trained weights, and hyperparameters.\footnote{\url{https://github.com/observer4599/explanation-regularization-in-bnn}}
\end{abstract}

\keywords{Explainable Artificial Intelligence \and Deep Learning \and Bayesian Neural Networks}

\section{Introduction}
\Acp{nn} have in recent years shown high performance and been successful in many applications~\citep{Goodfellow_DeepLearning_2016,Silver_generalreinforcementlearning_2018,Esteva_guidedeeplearning_2019,Kiran_DeepReinforcementLearning_2022}. However, they can overfit on spurious features in training datasets and lose the ability to generalize~\citep{Szegedy_Intriguingpropertiesneural_2014,Lapuschkin_UnmaskingCleverHans_2019}. Furthermore, we understand how they work computationally, but are unable to extract high-level insights that make humans understand and trust them~\citep{Arrieta_ExplainableArtificialIntelligence_2020}.

\begin{figure}[t]
    \centering
    \includegraphics[width=\linewidth]{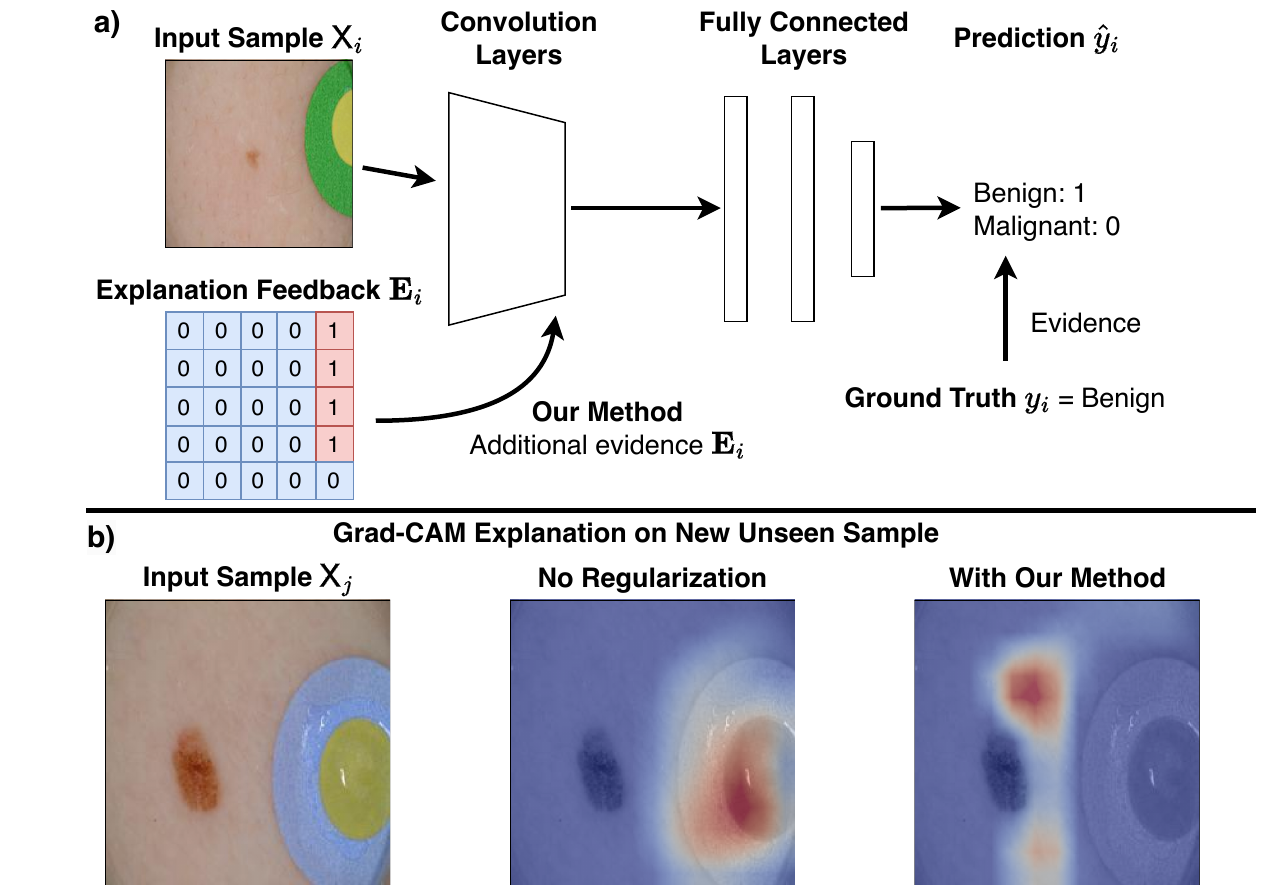}
    \caption{\textbf{Method Overview.} a) During training, a \ac{nn} gets an input sample \(\tX_i\in\R^{(w\times h\times c)}\) from the training dataset and tries to match the prediction \(\hat{y}_i\) with the ground truth label \(y_i\). Our method provides the \ac{nn} with additional evidence in the form of explanation feedback \(\mE_i\in\{0,1\}^{(w\times h)}\). A value of \(1\) in \(\mE_i\) indicates a region in the input space as irrelevant to the prediction, while \(0\) indicates that we do not have any concern. The explanation feedback is used to regularize the model's focus to give correct explanation and add domain knowledge. b) A new input sample \(\tX_j\) from the test dataset is fed to the model and an explanation is generated. Without explanation regularization, the \ac{nn} uses the patch to make the prediction. With our method, the \ac{nn} no longer looks at the patch in the image. The skin images are from the \ac{isic} dataset~\citep{Codella_SkinLesionAnalysis_2019,Tschandl_HAM10000DatasetLarge_2018,Rieger_InterpretationsareUseful_2020}.}
    \label{fig:method-overview}
\end{figure}

To prevent overfitting, we use regularization techniques like weight regularization, dropout, early stopping, and explanation regularization~\citep{Ross_RightRightReasons_2017}. A probabilistic approach to regularizing \acp{nn} is to leverage Bayesian inference~\citep{Blundell_WeightUncertaintyNeural_2015,Jospin_HandsBayesianNeural_2022}. In Bayesian \acp{nn}, we find the posterior distribution on weights rather than point estimates. To find the posterior distribution, we define a prior distribution on weights that moves them towards our preferred choices. As the amount of data increases, the prior weighs less~\citep{Blundell_WeightUncertaintyNeural_2015,Prince_UnderstandingDeepLearning_2023}. Although Bayesian inference gives us well-calibrated uncertainty estimates, this principled way to regularize \acp{nn} is incompatible with newer methods that regularize explanations. Explanation regularization came as a response to the need of explainable \acp{nn}~\citep{Ross_RightRightReasons_2017,Teso_ExplanatoryInteractiveMachine_2019,Rieger_InterpretationsareUseful_2020}. In explanation regularization, we have annotated masks that we refer to as \textit{explanation feedback}. They indicate areas in the input space irrelevant for predictions, which is seen in \cref{fig:method-overview}. Furthermore, Bayesian inference regularizes the model via prior on weights. However, it is unable to say anything regarding the input space. In contrast, explanation regularization enables us to add domain knowledge in the input space to regularize \acp{nn}' explanations, in the form of saliency maps. The ability to add domain knowledge in the input space, in turn, can make the models focus on the right features.

Our method provides a way to regularize explanations that is compatible with Bayesian \acp{cnn}. By merging Bayesian inference and our explanation regularization method, we introduce \acp{nn} with correctly calibrated uncertainty through a principled way and correct explanations that previous approaches have not been able to provide. Experimentally, we demonstrate that our method makes models perform better when they overfit to spurious features that a user can indicate in the input space. Furthermore, it can improve model performance when the model is uncertain on what to look at. We also show that our approach is more versatile than augmenting training data with samples where spurious features are masked.

To summarize: 1) we propose a new explanation regularization method compatible with Bayesian \acp{cnn} that provides well calibrated uncertainty estimate in a principled way. 2) We test our method on four different datasets with and without spurious features. 3) Experiments demonstrate that our method makes models perform better when they overfit to spurious features or are uncertain about which parts of the input to focus on.

\section{Background}
We introduce the background on Bayesian \ac{nn}~\citep{Prince_UnderstandingDeepLearning_2023,Murphy_ProbabilisticMachineLearning_2023} and the \ac{lrt}~\citep{Kingma_VariationalDropoutLocal_2015} that our method relies on. The loss function introduced in this section will be used in \cref{sec:method}.

\subsection{Bayesian Neural Network}\label{sec:background_bnn}
In \acp{nn}, we learn the weights \(\vw\) via maximum likelihood estimation. Given a dataset \(\data=\{(\vx_i, y_i)\}_{i=0}^N\) with \(N\) samples, we optimize the objective defined by \(\argmax_\vw \sum_{i=0}^N \log Pr(y_i|\vx_i,\vw)\) assuming that the samples are independent and identically distributed. There are several choices of regularization, one is to use the maximum a posteriori estimation defined by \(\argmax_\vw \sum_{i=0}^N \log Pr(y_i|\vx_i,\vw) + \log Pr(\vw)\), where \(Pr(\vw)\) moves the weights towards the choices we prefer to prevent overfitting. \(Pr(\vw)\) is referred to as the prior, and reflects our prior belief of what the weights should be before seeing the data. The prior imposes L1 or L2 regularization depending on if it is Laplace or Gaussian respectively.

Both maximum likelihood estimation and maximum a posterior estimation focus on finding point estimates of the weights. In Bayesian \acp{nn}, we represent weights as probability distributions and not as point estimates. To compute the full distribution \(Pr(\vw|\data)\) requires us to compute the integral \(\int Pr(\vy|\vx,\vw)Pr(\vw) d\vw\), which is infeasible. A way to solve this is to use \ac{vi}~\citep{DavidM.Blei_VariationalInferenceReview_2017} and minimize the \ac{kl} divergence \(\KL ( q_{\vtheta}(\vw) \Vert Pr(\vw|\data))\), where \(q_{\vtheta}(\vw)\) is the variational distribution and \(Pr(\vw|\data)\) is the posterior distribution~\citep{Blundell_WeightUncertaintyNeural_2015}. We cannot minimize the \ac{kl} divergence directly, but we can solve the optimization problem for a lower bound on the evidence that is independent of the distribution parameters \(\vtheta\). The lower bound is known as the \ac{elbo} and defined by
\begin{equation}\label{eq:elbo}
    \argmax_{\vtheta}  \EX_{\vw\sim q_{\vtheta}(\vw)}[\sum_{i=1}^N \log Pr(y_i|\vx_i,\vw)] - \KL ( q_{\vtheta}(\vw) \Vert Pr(\vw)).
\end{equation}
The objective maximizes the log likelihood of the data like in the maximum likelihood estimation. It is important to note that we are maximizing with respect to the distribution parameters \(\vtheta\) and not the weights themselves like in maximum likelihood estimation where we treated them as point estimates. The objective additionally minimizes the \ac{kl} divergence between the variational distribution and the prior distribution, moving the probability mass towards our choice of weights. The objective has to trade off between these two quantities, but as the amount of data increases, the likelihood term will weigh more.

To optimize \cref{eq:elbo}, we use stochastic gradient descent with the reparameterization trick~\citep{Kingma_AutoEncodingVariational_2014,Blundell_WeightUncertaintyNeural_2015}. We model the variational distribution with a fully factorized Gaussian distribution defined by \(q_\theta(\vw)=\prod_{i=0}^n \normal(w_i |\mu_i,\sigma_i^2)\) using the mean field approximation. To sample weights, we first sample noise \(\epsilon\sim\normal(\epsilon|0, 1)\), thereafter compute \(w_i = \mu_i + \sigma_i\epsilon\) for \(i\in\{1,2,\cdots,n\}\) independently. By using the reparameterization trick, we can update the parameters using backpropagation. The loss function we optimize in a Bayesian \ac{cnn} using minibatches is defined by
\begin{equation}\label{eq:sgd_elbo}
    L= \KL ( q_{\vtheta}(\vw) \Vert Pr(\vw)) - \frac{N}{M}\sum_{i=1}^M\sum_{j=1}^J\log Pr(\vy_i|\vx_i,\vw_j=\vmu+\vsigma\vepsilon_j),
\end{equation}
where \(M\) is the number of minibatches, \(N\) is the number of samples in our dataset and \(J\) the number of Monte Carlo samples. We use fully factorized Gaussians for both the variational distribution and the prior distribution so that the \ac{kl} divergence term can be solved in closed form~\citep{Kingma_AutoEncodingVariational_2014}.

\subsection{Local Reparameterization Trick}\label{ssec:background-lrt}
To reduce variance of \cref{eq:sgd_elbo}, \citet{Kingma_VariationalDropoutLocal_2015} propose the \acf{lrt}. Instead of sampling weights as in \cref{eq:sgd_elbo}, \ac{lrt} samples activations. Thus, the uncertainty is moved from weights that affect all samples to activations that is local and sample dependent. The \ac{lrt} loss function is defined by
\begin{equation}\label{eq:lrt_elbo}
    L^\text{LRT} = \KL ( q_{\vtheta}(\vw) \Vert Pr(\vw)) - \frac{N}{M}\sum_{i=1}^M\sum_{j=1}^J\log Pr(\vy_i|\va_{i,j}),
\end{equation}
where we sample activations \(\va\) rather than weights. We omit \(\vx_i\) in the condition as no extra information is added given that we know the activations. Do note that we do need to know \(\vx_i\) in the first place to compute the activations. We show how these activations are sampled in fully connected layers and in convolutional layers~\citep{Kingma_VariationalDropoutLocal_2015,Molchanov_VariationalDropoutSparsifies_2017}. 

\textbf{Fully Connected Layer.} Assume that the input to a layer is \(\vb\in\R^m\), to compute the activation, we compute the mean and variance of the activation defined by \(\delta = \sum_{i=1}^m b_iw_{i,j}\) and \(\gamma^2 = \sum_{i=1}^m b_i^2w_{i,j}^2\). Thus, the distribution on the activation is \(\normal(a|\delta,\gamma^2)\) and can be sampled as shown in \cref{sec:background_bnn}. 

\textbf{Convolutional Layer.} Assume that the input to a layer is \(\mB\in\R^{w'\times h'}\) and the weights \(\mW\) is also a matrix. We assume only a single feature map to simplify the calculations. The mean and variance are defined by \(\mdelta = \mB * \mW\) and \(\mgamma^2 = \mB^2 * \mW^2\) where \(*\) is the convolution operator and \((\cdot)^2\) is applied element-wise. The distribution on activations \(\mA\) is then \(\normal(\vect(\mA)|\vect(\mdelta),\vect(\mgamma^2))\) and the reparameterization trick can be used to sample activations.

\section{Related Work} \Ac{xai} aims to assist humans understand artificial intelligence systems, their strength and weaknesses, provide understanding of how they will perform in unknown situations~\citep{Gunning_DARPAsExplainableArtificial_2019}. Methods to understand machine learning models are often divided into interpretable models and post hoc explainability~\citep{Lipton_mythosmodelinterpretability_2018,Arrieta_ExplainableArtificialIntelligence_2020,Murphy_ProbabilisticMachineLearning_2023}. Our method goes under post hoc explainability methods that are applied to models after training. The method we propose is related to a line of work that corrects or prevents models to look at spurious features. As far as we know, \citet{Ross_RightRightReasons_2017} introduced the first method to correct and prevent models to look at spurious features in the context of \ac{xai}. To prevent models from learning spurious features, \citet{Ross_RightRightReasons_2017} regularizes the input gradient in area specified by an explanation feedback. That is, they minimize the \(\ell_2\) norm of the input gradient in the region that is specified to be irrelevant by the user. \citet{Liu_IncorporatingPriorsFeature_2019} use a similar approach to \citet{Ross_RightRightReasons_2017} in text classification to make a model focus less on certain words. Similarly, working on text, \citet{Du_LearningCredibleDeep_2019} encourages sparse importance values on irrelevant features and that the models should be uncertain when important features are removed. \citet{Rieger_InterpretationsareUseful_2020} regularize explanations leveraging the method contextual decomposition explanation penalization. This allows them to penalize both feature importance and interaction. \citet{Shao_RightBetterReasons_2021} regularize explanations using influence functions and show that it is better than using input gradients. \citet{Erion_Improvingperformancedeep_2021} regularizes explanations by specifying domain knowledge regarding how explanations should be before training. For example, the total variation between feature importance values for pixels in image data should be low. Like the abovementioned methods, \citet{Selvaraju_TakingHINTLeveraging_2019} propose a new loss function to align human feedback on important features and where models look. Common to all these approaches is that they modify the loss function by augmenting it with additional terms. This, however, makes it impossible to minimize \ac{elbo} as the loss function is modified and augmented with new terms. We instead introduce a simple approach levering \ac{lrt} to add explanation feedback to prevent models to look at irrelevant features and add domain knowledge.

Differently from previously mentioned methods, \citet{Schramowski_Makingdeepneural_2020,Teso_ExplanatoryInteractiveMachine_2019} propose a model agnostic approach to regularize explanations by augmenting the training dataset with counterexamples. These counterexamples are the same as the samples in the training dataset, but where spurious features have been modified. These modifications can be replacing spurious features with random noise or use feature values from other samples without spurious features. We show in the experiments that it is less effective than our approach since location dependent spurious features cannot be removed. Furthermore, sometimes background information can be a positive influence, but this method does not allow partial use of features by models. Lastly, creating counterexamples introduces out-of-distribution samples into the training dataset that can negatively affect training.

\begin{figure}[t]
    \centering
    \includegraphics[width=\linewidth]{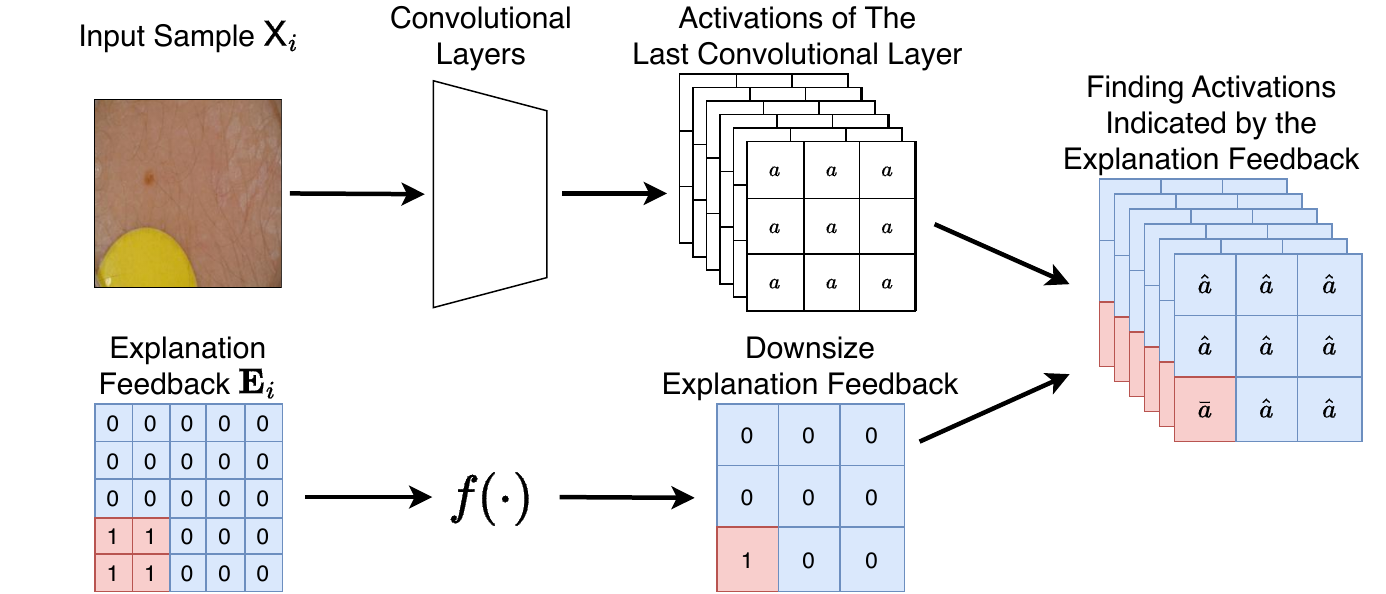}
    \caption{\textbf{Finding Activations.} Given an explanation feedback \(\mE_i\in\{0,1\}^{(w\times h)}\) for the sample \(\tX_i\in\R^{(w\times h\times c)}\), we find activations to add the explanation feedback. A value of \(1\) in \(\mE_i\) indicates irrelevant regions in the input. A value of \(0\) denotes features that no preference is given. First, \(\mE_i\) is downsized to the size of feature maps of the last convolutional layer using the function \(f(\cdot)\). Afterward, since the height and widths are the same, we simply overlay the explanation feedback with the feature maps to find activations to target. Specifically, we inject this information via the likelihood term of \cref{eq:lrt_elbo}. The skin image is from the \ac{isic} dataset~\citep{Codella_SkinLesionAnalysis_2019,Tschandl_HAM10000DatasetLarge_2018,Rieger_InterpretationsareUseful_2020}.}
    \label{fig:explanation-feedback}
\end{figure}

\section{Method}\label{sec:method}
We detail our method by first setting up the model and dataset assumptions. Afterward, we detail how to regularize explanations in Bayesian \ac{cnn} using our method. We assume that we have a Bayesian \ac{cnn} represented by \(Pr(y|\tX,\vw)\). Furthermore, we assume access to a dataset \(\data=\{(\tX_i,y_i,\mE_i)\}_{i=1}^N\) where \(\tX_i\in\R^{(w\times h\times c)}\) is an input image and \(y_i\in\Y\) is a target label. \(\Y\) is the set of real numbers if it is a regression task or a set of class labels for classification. \(\mE_i\in\{0,1\}^{(w\times h)}\) is an explanation feedback. A value of \(1\) in \(\mE_i\) indicates an area of \(\tX_i\) where the \ac{nn} should not focus on when predicting \(\hat{y}_i\). A value of \(0\) points at an area where no feedback is given, that is, it does not matter what the model does in that region.

We showed in \cref{ssec:background-lrt} that training a Bayesian \ac{cnn} with \ac{lrt} amounts to minimize \cref{eq:lrt_elbo}. To regularize explanations implies regularizing the input gradients~\citep{Ross_RightRightReasons_2017} or some other quantity~\citep{Rieger_InterpretationsareUseful_2020,Selvaraju_TakingHINTLeveraging_2019}. But to regularize input gradient without changing the objective, we need to know the distribution on input gradients, which we do not know. Instead, we leverage activation outputted from convolutional layers to incorporate the explanation feedback to regularize explanations. To show how our method works, we take the objective in \cref{eq:lrt_elbo} and show how the likelihood term is computed to incorporate explanation feedback.

We incorporate the explanation feedback via the last convolutional layer in a Bayesian \ac{cnn}. We downsize the explanation feedback to the size of the activation produced by the last convolutional layer, as seen in \cref{fig:explanation-feedback} using a function \(f(\cdot)\). In practice, the function is implemented using \texttt{torch.nn.AdaptiveMaxPool2d}~\citep{Ansel_PyTorch2Faster_2024}. Then we set the evidence of activation overlapping with \(1\)'s in the explanation feedback to \(0\). We denote those activations that the explanation feedback indicates are unimportant as \(\bar{\va}\), while the rest of the activations in the network as \(\hat{\va}\). When we refer to all activations in the network, we simply write \(\va\). The log likelihood term with explanation feedback added is defined by
\begin{equation}\label{eq:likelihood-exp}
\begin{split}
    \log Pr(\vy_i,\ve_i|\va_{i},\vx_{i}) &= \log Pr(\vy_i,\bar{\va}_i=\vzero|\vx_{i}) \\
    &=\underbrace{\log Pr(\vy_i|\bar{\va}_i=\vzero,\vx_{i})}_{\text{Correct Prediction}} + \underbrace{\log Pr(\bar{\va}_i=\vzero|\vx_i)}_{\text{Correct Explanation}}.
\end{split}
\end{equation}
Because the size of the explanation feedback is larger than the prediction output, we introduce a hyperparameter \(\lambda\) to lower the importance of \(\log Pr(\bar{\va}_i=\vzero|\hat{\va}_i)\) in \cref{eq:likelihood-exp} and set it to \(\lambda \ll 1\). Note that we still minimize \cref{eq:lrt_elbo} but add explanation feedback using activations as seen in \cref{fig:explanation-feedback} via the likelihood term as shown in \cref{eq:likelihood-exp}.

\section{Experiments}
We first detail our experimental setup, including the datasets used, model architectures, and additional details. Afterward, we show how our model improves the predictive performance while minimizing the models' focus on spurious features.

\subsection{Experimental Setup}
To test the performance of our method, we use four different datasets. All of the datasets except for the \ac{isic} skin cancer dataset were downloaded via \texttt{torchvision.datasets}~\citep{Ansel_PyTorch2Faster_2024}.

\begin{figure}[t]
\centering
\begin{subfigure}[t]{0.48\textwidth}
    \includegraphics[width=\linewidth]{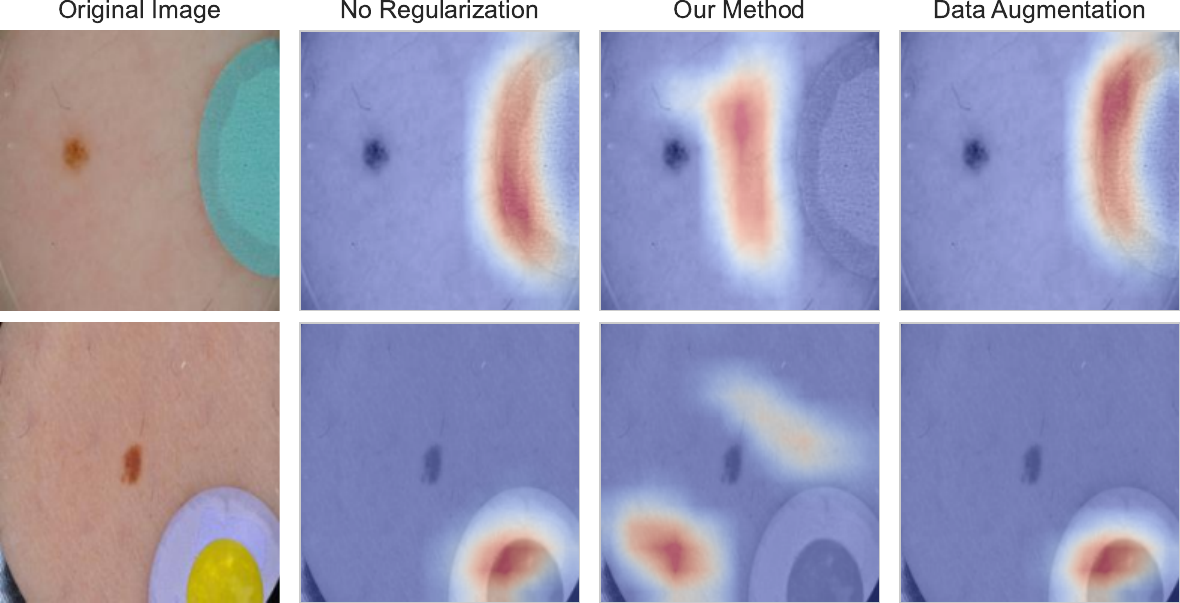}
    \caption{\textbf{ISIC.} Our method removes the focus on patches that the data augmentation approach is unable to.}
    \label{fig:isic-model-focus}
\end{subfigure}
\hfill
\begin{subfigure}[t]{0.48\textwidth}
    \includegraphics[width=\linewidth]{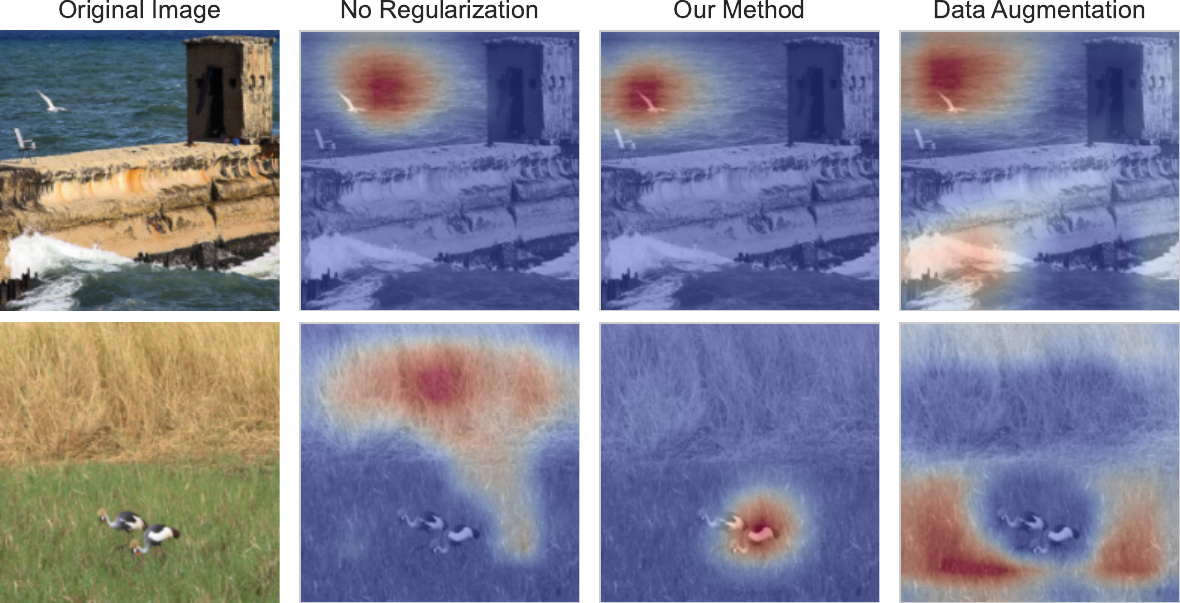}
    \caption{\textbf{SBD.} Our method makes the saliency maps more focused and concentrated.}
    \label{fig:sbd-model-focus}
\end{subfigure}
\hfill
\begin{subfigure}[t]{0.48\textwidth}
    \includegraphics[width=\linewidth]{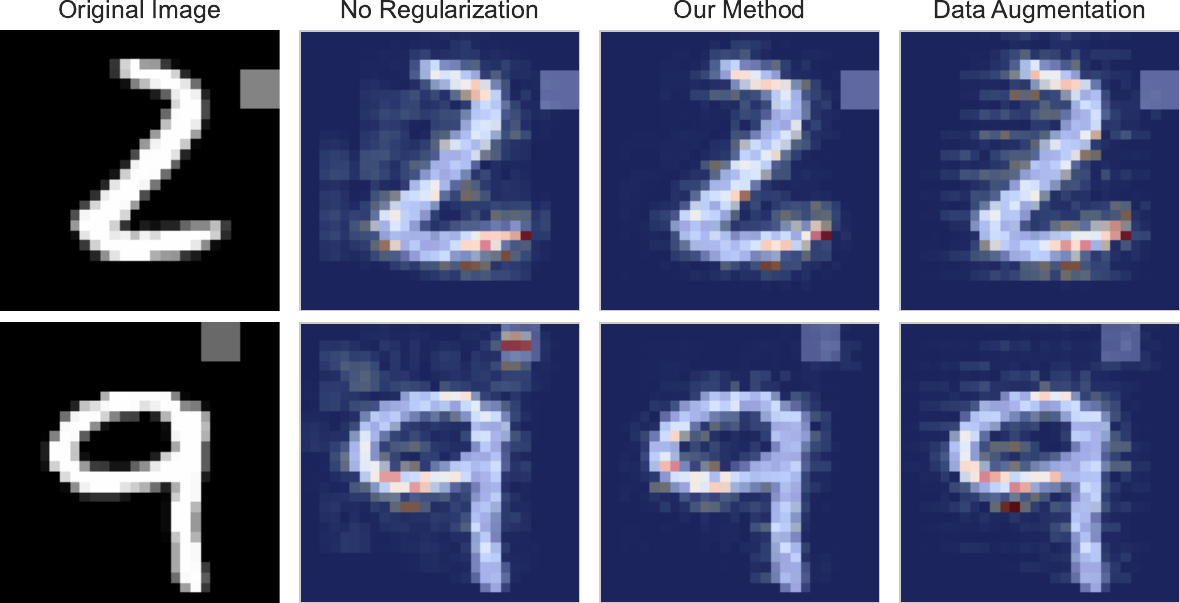}
    \caption{\textbf{Decoy MNIST Color.} Both our method and data augmentation can remove focus on decoys. Our method makes the saliency maps more focused.}
    \label{fig:decoy-mnist-model-focus}
\end{subfigure}
\hfill
\begin{subfigure}[t]{0.48\textwidth}
    \includegraphics[width=\linewidth]{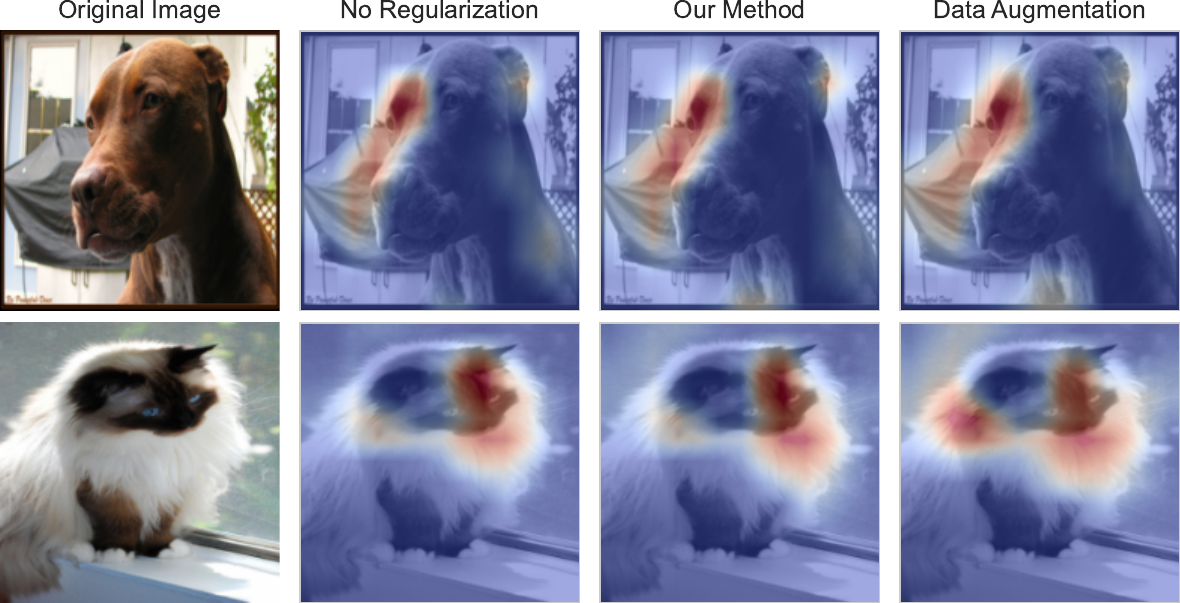}
    \caption{\textbf{Oxford-IIIT-Pet.} When no performance gain can be made, the saliency maps are similar.}
    \label{fig:oxford-iiit-pet-model-focus}
\end{subfigure}
\caption{Examples of saliency maps on samples randomly drawn from the test dataset. More examples can be found in the link given on the first page.}
\label{fig:figures}
\end{figure}

\textbf{Datasets.} We create two versions of Decoy MNIST~\citep{Ross_RightRightReasons_2017} which builds on The MNIST database of handwritten digits~\citep{LeCun_MNISThandwrittendigit_2010}. The MNIST dataset consists of black and white images of digits from 0 to 9. The Decoy MNIST dataset adds decoys at the corners and sides of input samples as seen in \cref{fig:decoy-mnist-model-focus}. In the first version that we name ``color'', the grayscale of decoys in the training has pixel intensity \(255 - 25k_i\) where \(k_i\) is the class label. In the test dataset, \(k_i\) is randomly sampled from the set of class labels. The location of the decoy is randomly placed both in the training and test dataset. In the other version called ``location'', the location of the decoys follows the class label in the training dataset but is random in the testing dataset. The grayscale intensity is randomly drawn both for the training and testing datasets. The \ac{isic} dataset is a dataset for skin cancer diagnosis~\citep{Codella_SkinLesionAnalysis_2019,Tschandl_HAM10000DatasetLarge_2018}. We utilize only two classes, benign and malignant. We increase the importance of the malignant class in the loss because the dataset is heavily imbalanced. The version of \ac{isic} dataset we use is curated by using code from \citet{Rieger_InterpretationsareUseful_2020}. The explanation feedback we used is also from \citet{Rieger_InterpretationsareUseful_2020}. Oxford-IIIT-Pet~\citep{Parkhi_Catsdogs_2012} consists of cat and dog images with 37 different classes of different cat and dog breeds. The \ac{sbd}~\citep{Hariharan_Semanticcontoursinverse_2011} dataset consists of images from the PASCAL VOC 2011 dataset~\citep{Everingham_PASCALVisualObject_}. For the \ac{sbd}, we use a subset of classes: bird, bus, cat, dog, horse by following \citet{Schramowski_Makingdeepneural_2020}. We only use samples where one and only one of these classes appears.

\textbf{Models.} We use the LeNet-5\footnote{\url{https://pytorch.org/tutorials/beginner/introyt/introyt1_tutorial.html\#pytorch-models}}~\citep{LeCun_Gradientbasedlearning_1998} model for the decoy MNIST datasets and AlexNet~\citep{Krizhevsky_ImageNetClassificationDeep_2012} for the other datasets. We load pretrained weights from PyTorch for AlexNet\footnote{\url{https://pytorch.org/hub/pytorch_vision_alexnet/}}.

\textbf{Software and Hardware.} We used PyTorch Lightning to do the experiments~\citep{Falcon_PyTorchLightning_2024}. The experiments ran on a MacBook Pro 2023 with Apple M2 Max chip and 64 GB RAM. We used the MPS backend for GPU accelerated training. The metrics we compute are calculated using scikit-learn~\citep{Pedregosa_ScikitlearnMachine_2011}. The saliency maps are created using Captum~\citep{Kokhlikyan_Captumunifiedgeneric_2020}.

\subsection{Predictive Performance}
We compare the predictive performance of Bayesian \acp{cnn} without any feedback, using data argumentation with counterexamples~\citep{Schramowski_Makingdeepneural_2020,Teso_ExplanatoryInteractiveMachine_2019}, and the method outlined above. The data augmentation approach is, as far as we know, the only approach compatible with Bayesian \acp{cnn} because it is model agnostic. For this approach, we first replace a region specified to be irrelevant by the explanation feedback with noise sampled from a uniform distribution on the interval \([0,1)\) and afterward, we preprocess the images with standardization. We only use \(70\%\) of the explanation feedback available, since regularizing all training samples negatively impacts our method in some instances.

We observed during the experiments that there are no performance gains when we apply our method to models that are not focusing on spurious features or when models are not uncertain. That is, if we initialized weights with small variance we could not see performance gain in the datasets without spurious features because pretrained AlexNet weights from PyTorch are already near optimal for the model architecture. Instead, we want to demonstrate our method under the conditions that there are spurious features or when the models are uncertain by initializing with larger variance and compare it to the data augmentation method. \cref{tab:performance-i,tab:performance-ii} indicate that our method can improve model performance when models have overfitted to spurious features or the model is uncertain. The sample standard deviations shown in \cref{tab:performance-i,tab:performance-ii} are computed by training three models using a 3-fold cross-validation and testing the three models on an independent test dataset.

For the data augmentation method, we see that the method can affect results negatively when the models are not overfitting to spurious features but still uncertain. This indicates that background information can be useful, but since the information is removed entirely, the models cannot take advantage of it. While our method tries to tell the models where to not look, we do not remove the information entirely and can use the hyperparameter \(\lambda\) to balance this aspect.

\begin{table}[t]
\centering
\caption{Predictive performance across four datasets with different variations of Decoy MNIST and \ac{isic}. For datasets with more than two classes, we compute macro-averaged F1 score.}
\begin{adjustbox}{width=1\textwidth}
\begin{tabular}{@{}
l S[table-format = 1.3(3)]S[table-format = 1.3(3)] S[table-format = 1.3(3)]S[table-format = 1.3(3)] S[table-format = 1.3(3)]S[table-format = 1.3(3)]@{}}
\toprule
Dataset & \multicolumn{2}{c}{No Regularization} & \multicolumn{2}{c}{Our Method} & \multicolumn{2}{c}{Data Augmentation} \\
& \multicolumn{1}{r}{Balanced Accuracy \(\uparrow\)} & \multicolumn{1}{r}{F1 \(\uparrow\)} & \multicolumn{1}{r}{Balanced Accuracy \(\uparrow\)} & \multicolumn{1}{r}{F1 \(\uparrow\)} & \multicolumn{1}{r}{Balanced Accuracy \(\uparrow\)} & \multicolumn{1}{r}{F1 \(\uparrow\)} \\ \cmidrule(lr){1-1}\cmidrule(lr){2-3}\cmidrule(lr){4-5}\cmidrule(lr){6-7}
Decoy MNIST Color & .966(003) & .966(003) & .977(001) & .977(001) & \B .978(001) & \B .978(001) \\
Decoy MNIST Position & .523(025) & .524(025) & \B .810(012) & \B .808(012) & .594(028) & .594(028) \\
\Ac{isic} & 0.836(0002) & 0.486(0026) & 0.832(0002) & \B 0.494(0014) & \B 0.838(0004) & 0.458(0020) \\
\Ac{isic} (No Patch Data) & 0.744(0012) & 0.486(0022) & \B 0.752(0006) & \B 0.497(0013) & 0.728(0016) & 0.461(0022) \\
Oxford-IIIT-Pet & .582(002) & .579(003) & \B .583(007) & \B .580(006) & .545(006) & .537(006) \\
\Ac{sbd} & .600(007) & .558(023) & \B .687(009) & \B .661(009) & .557(019) & .498(028) \\
\bottomrule
\end{tabular}
\end{adjustbox}
\label{tab:performance-i}
\end{table}

\begin{table}[t]
\centering
\caption{For dataset with more than two classes, we compute one-vs-rest to get the AUC scores. To compute overlap, we use input gradient for Decoy MNIST and Grad-CAM for the rest of the datasets. Some entries are missing standard deviation, since it is less than \(0.001\).}
\begin{adjustbox}{width=1\textwidth}
\begin{tabular}{@{}
l S[table-format = 3.3(3)]S[table-format = 3.3(3)] S[table-format = 3.3(3)]S[table-format = 3.3(3)] S[table-format = 3.3(3)]S[table-format = 3.3(3)]@{}}
\toprule
Dataset & \multicolumn{2}{c}{No Regularization} & \multicolumn{2}{c}{Our Method} & \multicolumn{2}{c}{Data Augmentation} \\
& \multicolumn{1}{r}{AUC \(\uparrow\)} & \multicolumn{1}{r}{Overlap \(\downarrow\)} & \multicolumn{1}{r}{AUC \(\uparrow\)} & \multicolumn{1}{r}{Overlap \(\downarrow\)} & \multicolumn{1}{r}{AUC \(\uparrow\)} & \multicolumn{1}{r}{Overlap \(\downarrow\)} \\ \cmidrule(lr){1-1}\cmidrule(lr){2-3}\cmidrule(lr){4-5}\cmidrule(lr){6-7}
Decoy MNIST Color & .999(000) & .028(004) & \B 1.000(000) & .009(000) & \B 1.000(000) & \B .007(000) \\
Decoy MNIST Position & .880(011) & .033(001) & \B .977(003) & \B .015(000) & .887(009) & .050(002) \\
\Ac{isic} & \B 0.921(0001) & 0.229(0008) & 0.920(0.001) & \B 0.002(0) & 0.919(0002) & 0.252(0013) \\
\Ac{isic} (No Patch Data) & 0.849(0003) & {n/a} & \B 0.850(0002) & {n/a} & \B 0.850(0002) & {n/a} \\
Oxford-IIIT-Pet & \B .968(001) & \B .121(002) & .967(001) & .176(004) & .962(000) & .127(003) \\
\Ac{sbd} & .866(004) & .538(021) & \B .892(004) & \B .515(017) & .827(008) & .541(008) \\
\bottomrule
\end{tabular}
\end{adjustbox}
\label{tab:performance-ii}
\end{table}

\subsection{Model Focus}
\cref{tab:performance-ii} demonstrate that our method is good at removing the models' focus on spurious features. The overlap is computed by calculating how much importance is on the area the explanation feedbacks indicate as unimportant divided by the total amount of importance across the entire image. To do the overlap calculation, we use input gradient~\citep{Simonyan_DeepConvolutionalNetworks_2014} for the MNIST dataset and we used Grad-CAM~\citep{Selvaraju_GradCAMVisual_2017} for the rest of the datasets. \cref{fig:decoy-mnist-model-focus,fig:isic-model-focus,fig:sbd-model-focus} show that our method can guide models away from spurious features and focus on what is important. For \ac{isic}, data augmentation replace irrelevant regions with random noise but seems to be unable to make the models not look at patches. This indicates that when the location of features matter and not only their appearance, then counterexamples are unable to change model focus.

\section{Conclusion and Discussion}
We have introduced a new explanation regularization methods that is compatible with the Bayesian formalism. Our focus has been to introduce a method that can be used with Bayesian \acp{cnn} and not compete with methods trying to improve model focus on regular \acp{nn}. Beyond this, we provide the opportunity to add domain knowledge in the input space. The experiments across four datasets show that our method can improve predictive performance of Bayesian \acp{cnn} when they overfit to spurious features or are uncertain where to focus. Moreover, we can remove focus on spurious features, no matter if it is because of appearance or their location.

While our method is simple, it has limitations. Like other explanation regularization methods, our method requires human labor to specify explanation feedback that can be labor-intensive. In the future, intelligent ways to obtain explanation feedback should be considered. We regularize across all channels in a region in the convolutional layers, which can potentially be undesirable. We should for future work investigate adaptive methods to intelligently select specific filters to regularize.

\section*{Acknowledgements}
The underlying template for this paper is made by \citet{Kour_Latexstyletemplate_2020}, licensed under the MIT License (\url{https://github.com/kourgeorge/arxiv-style/blob/master/License.txt}). The mathematical notation we use follows \citet{Goodfellow_DeepLearning_2016} closely.

\bibliographystyle{unsrtnat}
\bibliography{references}

\end{document}